\documentclass[a4paper,12pt,oneside]{amsart}
\usepackage{authblk}

\usepackage{geometry}
\usepackage{epstopdf}
\usepackage{mathtools}
\usepackage[cmtip,all]{xy}

\usepackage{graphicx}
\usepackage{amssymb}
\usepackage{amsfonts}
\usepackage{amsmath}
\usepackage{mathtools}
\usepackage{epstopdf}
\usepackage{color}
\usepackage[all,cmtip]{xy}
\usepackage{array}
\usepackage{tikz-cd}
\usepackage{tabularx}
\usepackage{fancyhdr}

\usepackage{lipsum}

\usepackage{bm,dsfont}
\usepackage{multirow}
\usepackage{enumitem}
\usepackage{color}
\definecolor{red}{rgb}{0.70,0.13,0.13}
\definecolor{green}{rgb}{0.13,0.55,0.13}
\definecolor{blue}{rgb}{0.25, 0.41, 0.88}
\usepackage{hyperref}
\hypersetup{
colorlinks = true,
linkcolor = blue,
citecolor = green,
urlcolor = blue}

\newcommand{\id}{\mathrm{id}}

\newcommand{\dsE}{\mathbb{E}}

\newcommand{\dsR}{\mathbb{R}}

\newcommand{\scA}{\mathcal{A}}

\newcommand{\scC}{\mathcal{C}}
\newcommand{\scD}{\mathcal{D}}

\newcommand{\scF}{\mathcal{F}}

\newcommand{\scL}{\mathcal{L}}

\newcommand{\Obj}{\mathcal{O}bj}
\newcommand{\Hom}{\mathcal{H}om}

\newcommand{\eq}[1]{\begin{equation}#1\end{equation}}
\newcommand{\eqs}[1]{\begin{equation}\begin{split}#1\end{split}\end{equation}}
\newcommand{\eqnref}[1]{Eq.\,\eqref{#1}}
\newcommand{\figref}[1]{Fig.\,\ref{#1}}

\usepackage{times} 

\usepackage[english]{babel}
\usepackage[numbers,sort&compress]{natbib}

\theoremstyle{plain}

\newtheorem{thm}{Theorem}[section]

\newtheorem*{prop*}{Proposition}

\newtheorem{prop-defi}[thm]{Proposition-Definition}
\newtheorem{thm-defi}[thm]{Theorem-Definition}
\newtheorem{lem-defi}[thm]{lema-Definition}

\newtheorem{exam}[thm]{Example}

\newenvironment{itemize*}%
  {\vspace{-5pt}\begin{itemize}%
    \setlength{\itemsep}{0pt}%
    \setlength{\parskip}{0pt}}%
  {\end{itemize}\vspace{-5pt}}
 \newenvironment{enumerate*}%
  {\vspace{-5pt}\begin{enumerate}[label={(\arabic*)}]%
    \setlength{\itemsep}{0pt}%
    \setlength{\parskip}{0pt}}%
  {\end{enumerate}\vspace{-5pt}}
\usepackage{tikz-cd}

\begin{document}

\title{Categorical Representation Learning: Morphism is All You Need}
\maketitle

\begin{center}
\author{Artan Sheshmani and Yizhuang You}{Artan Sheshmani${}^{1,2,3, 5}$ and Yizhuang You${}^{4,5}$ with an Appendix by Ahmadreza Azizi${}^{5}$}
\end{center}
\address{${}^1$  Center for Mathematical Sciences and\\ Applications, Harvard University, Department of Mathematics, 20 Garden Street, Cambridge, MA, 02139}
\address{${}^2$ Institut for Matematik, Ny Munkegade 118
Building 1530, DK-8000 Aarhus C, Denmark}
\address{${}^3$ National Research University Higher School of Economics, Russian Federation, Laboratory of Mirror Symmetry, NRU HSE, 6 Usacheva str.,Moscow, Russia, 119048}
\address{${}^4$ University of California Sandiego, Department of Physics,Condenced matter group, UC San Diego 9500 Gilman Dr. La Jolla, CA 92093}
\address{${}^5$ QGNai INC. (Quantum Geometric networks for artificial intelligence), 83 Cambridge Parkway, Unit W806. Cambridge, MA, 02142}
\date{\today}

\begin{abstract}
We provide a construction for categorical representation learning and introduce the foundations of ''\textit{categorifier}". The central theme in \emph{representation learning} is the idea of {\bf everything to vector}. Every object in a dataset $\mathcal{S}$ can be represented as a vector in $\mathbb{R}^n$ by an \emph{encoding map} $E: \Obj(\mathcal{S})\to\mathbb{R}^n$. More importantly, every morphism can be represented as a matrix $E: \Hom(\mathcal{S})\to\mathbb{R}^{n}_{n}$. The encoding map $E$ is generally modeled by a \emph{deep neural network}. The goal of representation learning is to design appropriate tasks on the dataset to train the encoding map (assuming that an encoding is optimal if it universally optimizes the performance on various tasks). However, the latter is still a \emph{set-theoretic} approach. The goal of the current article is to promote the representation learning to a new level via a \emph{category-theoretic} approach. As a proof of concept, we provide an example of a text translator equipped with our technology, showing that our categorical learning model outperforms the current deep learning models by 17 times. The content of the current article is part of the recent US patent proposal submitted by the authors (patent application number: 63110906). 
\smallskip



\end{abstract}
\tableofcontents

\section{Introduction}

\begin{figure}[htbp]
\begin{center}
\includegraphics[width=0.7\columnwidth]{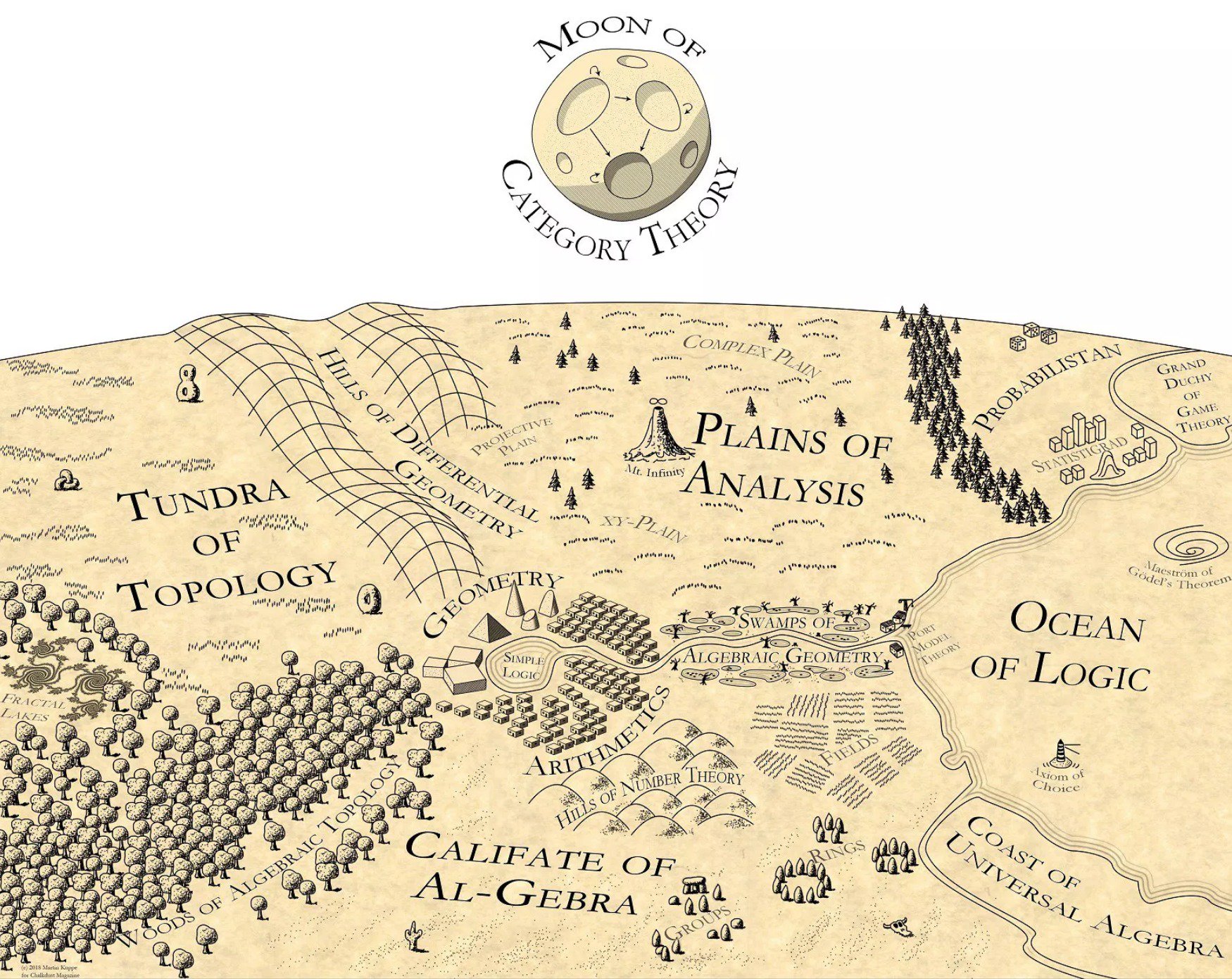}
\caption{Martin Kuppe’s map of the mathematical landscape (with permission to reuse).}
\label{fig: map}
\end{center}
\end{figure}

The rise of category theory was a great revolution of mathematics in the 20th century. Martin Kuppe once created a wonderful map of the mathematical landscape (see \figref{fig: map}) in which category theory hovers high above the ground, providing a sweeping vista of the terrain. It enables us to see relationships between various fields that are otherwise imperceptible at ground level, attesting that seemingly unrelated areas of mathematics aren’t so different after all. A category describes a collection of {\it objects} together with the relations (called {\it morphisms}) between them. The concept of category has provided a unified template for different branches of mathematics. Category theory is expected to be suitable for modeling datasets with relational structures, such as semantic relations among words, phrases and sentences in language datasets, or social relations among people or organizations in social networks.

From the category theory perspective, {\it relationships are everything}. Objects must be defined and can only be defined through their interrelations. So far, the field of representation learning\cite{bengio2013representation,goodfellow2016deep} has remained focused on learning object encodings, following the idea of ``everything to vector''. More recently, graph neural network models\cite{scarselli2008graph,zhou2018graph} and self-attentive mechanisms\cite{vaswani2017attention} have begun to model relationships between objects, but they still did not place relationships in the first place. This article aims to develop a novel machine learning framework, called \textit{\textbf {categorical representation learning}}, that directly learns the {\it representation of relations} as feature matrices (or tensors). The learned relations will then be used in different tasks. For example, direct relations can be composed to reveal higher-order relations. Functor map between different categories can be learned by aligning the relations (morphisms). Relations can guide the algorithm to identify clusters of objects and to perform renormalization transformations to simplify the category.

\subsection{Overview}

The construction of the proposed architecture in the current article will contain three major steps:
\begin{itemize}
\item {\bf Step 1: Mine the categorical structure from data}. In this step the objective is to develop the categorical representation learning approach, which enables the machine to extract the representation of objects and morphisms from data. The approach developed in this step will be the foundation of the construction which provides the morphism representation to drive further applications.
\item {\bf Step 2: Align the categorical structures between datasets}. The objective in this step is to develop the functorial learning approach, which can establish a functor between categories based on the learned categorical representations by aligning the morphism representations. The technique developed in this step will find applications in unsupervised (or semi-supervised) translation.  
\item {\bf Step 3: Discover hierarchical structures with tensor categories}. The objective in this step is to combine the categorical representation learning and functorial learning in the setting of a tensor category to learn the tensor functor that fuses simple objects into composite objects. The approach here will enable the algorithm to perform renormalization group transformations on categorical dataset, which progressively simplifies the category structure. This will open up broad applications in classification and generation tasks of categories.
\end{itemize}

\section{Aknowledgement}
We would like to thank Max Tegmark, Ehsan Kamali Nezhad, Lindy Blackburn for valuable discussions. 
\section{Categories and categorical embedding}

A category $\scC$ consists of a set of objects, $\Obj(\scC)$ ,  and a set of morphisms $\Hom(\scC)$ between objects. Each morphism $f$ maping a source object $a$ to a target object $b$, is denoted as $f:a\to b$. The hom-class $\Hom(a,b)$ denotes the set of all morphisms from $a$ to $b$. Morphisms benefit from composition laws. The composition is a binary operation; $\circ: \Hom(a,b)\times\Hom(b,c)\to\Hom(a,c)$ defined for every three objects $a,b,c$. The composition of morphisms $f:a\to b$ and $g:b\to c$ is written as $g\circ f$. The composition map is associative, that is, $h\circ(g\circ f)=(h\circ g)\circ f$. Given a category $\mathcal{C}$, for every object $x\in \Obj(\mathcal{C})$ there exists an identity morphism $\id_x:x\to x$, such that for every morphism $f:a\to x$ and $g:x\to b$, one has $\id_x\circ f= f$ and $g\circ \id_x=g$, that is pre and post compositions of morphisms in the category with the identity morphism will leave them unchanged.

One can associate a vector space to a given category, as its representation space. For the latter one can think of such vector space as a category itself, and use a categorical functors as a map from a given category to the geometric space.

Take for instance a category $\scC$ in which every {\it object}, $a$ is mapped to a {\it vector} $v_a$ in an ambient vector space $R$, and every morphisms between two objects in $\scC$, $f: a\to b$, is mapped to a morphism between vectors $v_a,v_b$ in $R$. Since $R$ is a vector space, a morphism between two vectors is then realized in $R$ as a {\it matrix} $M_f: v_a\to v_b$. For our purposes, the object and morphism embeddings, discussed above, will be implemented by separate embedding layers in a neural network. The map $f$ is a morphism from object $a$ to object $b$, iff the embedding  matrix can transform the embedding vectors as $v_b= M_f v_a$ under matrix-vector multiplication. The composition of morphisms is then realized as matrix-matrix multiplications as $M_{g\circ f}=M_g M_f$, which is naturally associative. The identity morphism of an object $x\in \scC$ is then represented as the projection operator $M_{\id_x}=v_x v_x^\intercal /|v_x|^2$, which preserves the vector representation $v_x$ of the object. In this manner, the category structure is  then represented respectively by the object, and morphism embeddings in the feature space.

\begin{exam}
For a language dataset, each object is given as a \emph{word}, represented as a word vector, via the \textit{word-vector} mapping functor. Each morphism is a \emph{relation} between two words, represented by a matrix. For instance
\eqs{
\texttt{bright}\xrightarrow{\texttt{antonym}}\texttt{dark}&: v_{\texttt{dark}}=M_{\texttt{antonym}} v_{\texttt{bright}},\\
\texttt{socks}\xrightarrow{\texttt{inside}}\texttt{shoes}&: v_{\texttt{shoes}}=M_{\texttt{inside}} v_{\texttt{socks}}.}
Given any pair of word vectors $v_a,v_b$, the hom-class $\Hom(v_a,v_b)$ denotes the set of all matrices $M$ which transform $v_a$ to $v_b$, that is; $v_b=M v_a$. 
\end{exam}

\subsubsection{Fuzzy Morphisms}

In machine learning, the relation between objects may not be strict, that is; \,$M_f v_a$ will not match $v_b$ precisely, but only align with $v_b$ with a high probability. To account for the fuzziness of relations then, the statement of $f:a\to b$ should be replaced by the probability of $f$ belonging to the class $\Hom(a,b)$. It can be written as
\eq{P(f\in\Hom(a,b))\equiv P(a\xrightarrow{f} b)\propto \exp z(a\xrightarrow{f} b),}
which is parametrized by the logit $z(a\xrightarrow{f} b)\in \dsR$. The logit is proposed to be modeled by
\eq{\label{eq: z(f:a->b)}z(a\xrightarrow{f} b)=v_b^\intercal M_f v_a,}
such that if $v_b$ and $M_f v_a$ align in similar directions, the logit will be positive, which results in a high probability for the morphism $f$ to connect $a$ to $b$. The objects $a$ and $b$ are said to be related (linked) if there exists at least one morphism connecting them. 

The linking probability is then proportional to the sum of the likelihood of each candidate morphism, $P(a\to b)\propto \sum_f \exp{z(a\xrightarrow{f} b)}$. The probability is normalized by considering the unlinking likelihood as the unit. Thus the binary probability $P(a\to b)$ can be modeled by the sigmoid of a logit $z(a\to b)$ that aggregates the contributions from all possible morphisms,
\eqs{\label{eq: P(a->b)}P(a\to b)&=\mathsf{sigmoid}(z(a\to b))\equiv \frac{e^{z(a\to b)}}{1+e^{z(a\to b)}},\\
z(a\to b)&=\log\sum_f \exp z(a\xrightarrow{f} b)=\log\sum_f \exp(v_b^\intercal M_f v_a),}
where it is assumed that each morphism contributes to the linking probability independently. More generally, the assumption of the independence of morphisms in determining the linking probability can be relaxed, such that the logits $z(a\xrightarrow{f} b)$ of different morphisms $f$ is aggregated by a generic nonlinear function:
\eq{z(a\to b)=F\Big(\bigoplus_f z(a\xrightarrow{f} b)\Big)=F\Big(\bigoplus_f v_a^\intercal M_f v_b\Big),}
where $F$ may be realized by a deep neural network, and $\bigoplus_f$ denotes the concatenation of logit contributions from different morphisms.

\subsubsection{Learning Embedding from Statistics}

The key idea of unsupervised representation learning is to develop feature representations from data statistics. In the categorical representation learning, the object and morphism embeddings are learned from the concurrence statistics, which corresponds to the probability $p(a,b)$ that a pair of objects $(a,b)$ occurs together in the same composite object. For example, two concurrent objects could stand for two elements in the same compound, or two words in the same sentence, or two people in the same organization. Concurrence does not happen for no reason. Assuming that two objects appearing together in the same structure can always be attributed to the fact that they are related by at least one type of relations, then the linking probability $P(a\to b)$ should be maximized for the observed concurrent pair $(a,b)$ in the dataset.

The negative sampling method can be adopted to increase the contrast. For each observed (positive) concurrent pair $(a,b)$, the object $b$ will be replaced by a random object $b'$, drawn from the negative sampling distribution $p_N(b')$ among all possible objects. The replaced pair $(a,b')$ will be considered to be an unrelated pair of objects. Therefore the training objective is to maximize the linking probability $P(a\to b)$ for the positive sample $(a,b)$ and also maximize the unlinking probability $P(a\nrightarrow b')=1-P(a\to b')$ for a small set of negative samples $(a,b')$,
\eq{\label{eq: Lneg}\scL=\mathop{\dsE}_{(a,b)\sim p(a,b)}\Big(\log P(a\to b)+\mathop{\dsE}_{b'\sim p_N(b')}\log(1-P(a\to b'))\Big).}
With the model of $P(a\to b)$ in \eqnref{eq: P(a->b)}, the object embedding $v_a$ and morphism embedding $M_f$ can be trained by maximizing the objective function $\scL$. The negative sampling distribution $p_N(b')$ can be engineered, but the most natural choice is to take the marginalized object distribution $p_N(b')=p(b')=\sum_a p(a,b')$. The theoretical optimum \cite{levy2014neural} is achieved with the following logit
\eq{z(a\to b)=\log\frac{p(a,b)}{p(a)p(b)}=\mathsf{PMI}(a,b),}
which learns the point-wise mutual information (PMI) between objects $a$ and $b$. If the objects were not related at all, their concurrence probability should factorize to the product of object frequencies, i.e.\,$p(a,b)=p(a)p(b)$, implying zero PMI. So a non-zero PMI indicates non-trivial relations among objects: a positive relation (PMI$>0$) enhances $p(a,b)$ while a negative relation (PMI$<0$) suppresses $p(a,b)$ relative to $p(a)p(b)$. By training the logit $z(a\to b)$ to approximate the PMI, the relations between objects are discovered and encoded as the feature matrix $M_f$ of morphisms.

\subsubsection{Scope of Concurrence}

It is noted that the definition of concurrence depends on the scope (or the context scale). For example, the concurrence of two words can be restricted in the scope of phrases or sentences or paragraphs. Different choices for the concurrence scope (say length of sentences, or phrases being considered) will affect the concurrent pair distribution $p(a,b)$, which then leads to different results for the object and morphism embeddings. This implies that objects may be related by different types of morphisms in different scopes. The categorical representation learning approach mentioned above can be applied to uncover the morphism embeddings in a scope-dependent manner. Eventually, the different object and morphism embeddings across various scopes can further be connected by the renormalization functor, which maps the category structure between different scales, detail of which will be elaborated in later sections. The scope-dependent categorical representation learning will form the basis for the development of unsupervised renormalization technique for categories with hierarchical structures, which will find broad applications in translation, classification, and generation tasks.

\subsubsection{Connection to Multi-Head Attention}

The multi-head attention mechanism\cite{vaswani2017attention} consists of two steps. The first step is a dynamic link prediction by the ``\textit{query-key matching}", that the probability to establish an attention link from $a$ to $b$ is given by $P(a\xrightarrow{f}b)\propto \exp z(a\xrightarrow{f}b)$ with $f$ labeling the attention head. The logit is computed from the inner product between the query vector , $Q_f v
_b$, and the key vector, $K_fv_a$, as \eq{z(a\xrightarrow{f}b)=v_b^\intercal Q_f^\intercal K_f v_a,}
which can be written in the form of \eqnref{eq: z(f:a->b)} if $M_f=Q_f^\intercal K_f$. The second step is the value propagation along the attention link (weighted by the linking probability).
Focusing on the first step of dynamic link prediction, the linking probability model in \eqnref{eq: P(a->b)} resembles the multi-head attention mechanism with the morphism embedding $M_f=Q_f^\intercal K_f$ given by the product of query and key matrices for each attention head. The proposal of the categorical representation learning is to keep the learned morphism matrix $M_f$ as encoding of relations in the feature space, which can be further used in other task applications.

The matrix $M_f$ can also be viewed as a metric in the feature space, which defines the inner product of object vectors. Different morphisms $f$ correspond to different metrics $M_f$, which distort the geometry of the feature space differently (bring object embeddings together or pushing them apart). When there are multiple relations in the action, the feature space is equipped with different metrics simultaneously, which resembles the idea of superposition of geometries in quantum gravity. If the single-head logit $z(a\xrightarrow{f} b)=v_b^\intercal M_f v_a$ is considered as an (negative) energy of two objects $a$ and $b$ embedded in a specific geometry, the aggregated multi-head logit  $z(a\to b)=\log\sum_f\exp z(a\xrightarrow{f} b)$ is analogous to the (negative) free energy that the objects will experience in the ensemble of fluctuating geometries.

\subsection{Aligning the Categorical Structures between Data sets}

\subsubsection{Tasks as Functors}

Functor is a fundamental concept in the category theory. It denotes the structure-preserving map between two categories. A functor $\scF$ from a source category $\scC$ to a target category $\scD$  is a mapping that associates to each object $a$ in $\scC$ an object $\scF(a)$ in $\scD$, and associates to each morphism $f:a\to b$ in $\scC$ a morphism $\scF(f):\scF(a)\to\scF(b)$ in $\scD$, such that $\scF(\id_a)=\id_{\scF(a)}$ for every object $a$ in $\scC$ and $\scF(g\circ f)=\scF(g)\circ\scF(f)$ for all morphisms $f:a\to b$ and $g:b\to c$ in $\scC$. The definition can be illustrated with the following commutative diagram.
\begin{equation}
\begin{tikzcd}
\mathcal{C} \arrow[d, "\mathcal{F}"] & a \arrow[r, "f"] \arrow[d, "\mathcal{F}"]  & b \arrow[d, "\mathcal{F}"] \\
\mathcal{D}                          & \mathcal{F}(a) \arrow[r, "\mathcal{F}(f)"] & \mathcal{F}(b)            
\end{tikzcd}
\end{equation}
A functor between two categories not only maps objects to objects, but also preserves their relations, which is essential in category theory. Many machine learning tasks can be generally formulated as functors between two data categories. For instance, machine translation is a functor between two language categories,  which not only maps words to words but also preserves the semantic relations between words. Image captioning is a functor from image to language categories, that transcribes objects in the image as well as their interrelations. 

In what follows, we discuss {\it Functorial learning} as an approach, providing the means for the machine to learn the functorial maps between categories in an unsupervised or semisupervised manner.

\subsubsection{Functorial Learning}

In functorial learning, each functor $\scF$ is represented by a transformation $V_\scF$ that transforms the vector embedding $v_a$ of each object $a$ in the source category to the vector embedding $v_{\scF(a)}$ of the corresponding object $\scF(a)$ in the target category
\eq{\label{eq: v=Vv} v_{\scF(a)}=V_{\scF} v_a,}
and also transforms the matrix embedding $M_f$ of each morphism in the source category to the matrix embedding $M_{\scF(f)}$ of the corresponding morphism $\scF(f)$ in the target category
\eq{\label{eq: MV=VM} M_{\scF(f)} V_{\scF} =V_{\scF}M_{f},}
represented by the following commutative diagram
\begin{center}
\begin{tikzcd}
\mathcal{C} \arrow[d, "\mathcal{F}"] & v_a \arrow[r, "M_f"] \arrow[d, "V_{\mathcal{F}}"]  & v_b \arrow[d, "V_{\mathcal{F}}"] \\
\mathcal{D}                          & V_{\mathcal{F}}v_a \arrow[r, "M_{\scF(f)}"] & V_{\mathcal{F}}v_b            
\end{tikzcd}
\\
\end{center}
In the case where the objects are embedded as unit vectors on a hypersphere, the functorial transformation $V_\scF$ will be represented by orthogonal matrices, whose inverses are simply given by the transpose matrices $V_\scF^\intercal$, which admit efficient implementation in the algorithm. The proposed representation for the functor automatically satisfies the functor axioms, as
\begin{equation}
\begin{split}
&M_{\scF(\id_a)}=v_{\scF(a)} v_{\scF(a)}^\intercal = V_\scF v_a v_a^\intercal V_{\scF}^\intercal=V_\scF M_{\id_a}V_{\scF}^\intercal,\notag\\
&M_{\scF(g\circ f)}=V_\scF M_{g\circ f}V_{\scF}^\intercal=V_\scF M_{g} M_{f}V_{\scF}^\intercal\\
&\phantom{M_{\scF(g\circ f)}}=V_\scF M_{g}V_\scF^\intercal V_\scF M_{f}V_\scF^\intercal=M_{\scF(g)}M_{\scF(f)}=M_{\scF(g)\circ\scF(f)}.\notag\\
\end{split}
\end{equation}
So as long as the optimal transformation $V_\scF$ can be found, our design will ensure that it parametrizes a legitimate functor between categories.

\subsubsection{Universal Structure Loss}

Now we explain how to find the optimal transformation $V_\scF$. The most important requirement for a functor is to preserve the morphisms between two categories. Hence we propose to train the functor by minimizing a loss function which ensures the structural compatibility in Equation \eqref{eq: MV=VM}. We denote the loss function by {\it universal structure loss},
\eq{\scL_\text{struc}=\sum_f \Vert M_{\scF(f)}V_{\scF} - V_{\scF}M_{f}\Vert^2,}
given the matrix representation $M_{f}, M_{\scF(f)}$ of morphisms in both the source and target categories, which were obtained from the categorical representation learning. The loss function is universal in the sense that it is independent of the specific task that the functor is trying to model. In this approach, the morphism embeddings are all that we need to drive the learning of functor transformation $V_\scF$. This is precisely in line with the spirit of the category theory: objects are illusions, they must be defined and can only be defined by morphisms. Therefore \begin{center}
``\textbf{\textit{Morphism is all you need!}}''
\end{center}.

\subsubsection{Alignment Loss}

However, in reality, the learned morphism matrices $M_f$ may not be of full rank. In such cases, the solution of the transformation $V_\scF$ is not unique, because arbitrary transformation within the null space of the morphism matrix can be composed with $V_\scF$ without affecting the structure lost $\scL_\text{struc}$. To overcome this difficulty, we propose to subsidize the structure loss with additional alignment loss over a few pair of aligned pairs of objects $(a,\scF(a))$,
\eq{\scL_\text{align}=\sum_{a\in\scA}\Vert v_{\scF(a)}- V_\scF v_a\Vert^2,}
where $\scA$ is only a subset of objects in the source category. This provides additional supervised signals to train $V_\scF$ by partially enforcing \eqnref{eq: v=Vv}. The total loss will be a weighted combination of the structure loss and the alignment loss
\eq{\scL=\scL_\text{struc}+\lambda \scL_\text{align}.}
By minimizing the total loss, $V_\scF$ will be trained, and the functor between categories can be established. The advantage of the categorical approach is that the structural loss already puts constrains on the parameters in $V_\scF$, so the effective parameter space for the alignment loss is reduced, such that the model can be trained with much less supervised samples when the category structure is rigid enough.

\subsection{Discovering Hierarchical Structures with Tensor Categories}

\subsubsection{Renormalization as Tensor Bifunctor}
Renormalization group plays an essential role in analyzing the hierarchical structures in physics and mathematics. It provides an efficient approach to extract the essential information of a system by progressively coarse graining the objects in the system. The categorical representation learning and functorial learning proposed in the previous two sections provide solid foundation to develop hierarchical renormalization approach for machine learning. This will enable the machine to summarize the feature representation of composite objects from that of simple objects.

The elementary step of a coarse graining procedure is to fuse two objects into one composite object (or ``higher'' object). Multiple objects can then be fused together in a pair-wise manner progressively. In category theory, the pair-wise fusion of objects is formulated as a tensor bifunctor $\otimes: \scC\times\scC \to \scC$. The tensor bifunctor maps each pair of objects $(a,b)$ to a composite object $a\otimes b$ and each pair of morphisms $(f,g)$ to a composite morphism $f\otimes g$ while preserving the morphisms between objects, as presented in the following diagram.
\begin{equation}
\begin{tikzcd}
\mathcal{C}\times\mathcal{C} \arrow[d] & {(a,b)} \arrow[r, "{(f,g)}"] \arrow[d] & {(f(a),g(b))} \arrow[d]  \\
\mathcal{C}                            & a\otimes b \arrow[r, "f\otimes g"]     & (f\otimes g)(a\otimes b)
\end{tikzcd}
\end{equation}
The category equipped with the tensor bifunctor is called a tensor category (or monoidal category). Objects and morphisms of different hierarchies are treated within the same framework systematically. This enables the algorithm to model multi-scale structure in the data set with the universal approach of categorical representation learning.

\subsubsection{Representing Tensor Bifunctor}

As a special case of general functors, the tensor bifunctor can be represented by a {\it fusion operator} $\Theta$, such that the object embeddings are fused by
\eq{v_{a\otimes b} = \Theta (v_a\otimes v_b),}
and the morphism embeddings are fused by
\eq{M_{f\otimes g}\Theta=\Theta (M_f\otimes M_g),}
following the general scheme in \eqnref{eq: v=Vv} and \eqnref{eq: MV=VM}. The tensor product of the vector and matrix representations are implemented as Kronecker products. $\Theta$ can be viewed as an operator which projects the tensor-product feature space back to the original feature space.

To simplify the construction, we will assume that the representation of the tensor bifunctor is strict in the feature space, meaning that the fusion is strictly associative without non-trivial natural isomorphisms,
\eq{\label{eq: associativity} v_{a\otimes b\otimes c}=\Theta(\Theta(v_a\otimes v_b)\otimes v_c)=\Theta(v_a\otimes \Theta(v_b\otimes v_c)).}
Such a strict representation will always be possible given a large enough feature space dimension, since every monoidal category is equivalent to a strict monoidal category. To impose the strict associativity in the learning algorithm, we propose to fuse objects in different orders, such that the machine will not develop any preference over a particular fusion tree and will learn to construct an associative fusion operator $\Theta$. With the fusion operator, we establish the embedding for all composite objects (and their morphisms) in the category given the embedding of fundamental objects (and their morphisms).

\subsubsection{Multi-Scale Categorical Representation Learning}

The tensor bifunctor learning can be combined with the categorical representation learning as an integrated learning scheme, which allows the algorithm to mine the category structure at multiple scales. If the data set has naturally-defined levels of scopes, one can learn the objects and morphism embeddings in different scopes. The objective is to learn the concurrence of objects within their scope according to the loss function \eqnref{eq: Lneg}. For each pair of objects $(a,b)$ drawn from the same scope, we want to maximize the linking probability $P(a\to b)$. For randomly sampled pairs of objects $(a,b')$, we want to maximize the unlinking probability $P(a\nrightarrow b')=1-P(a\to b')$. The linking probability $P(a\to b)$ is modeled by \eqnref{eq: P(a->b)}, based on the object and morphism embeddings. The vector embedding $v_a$ of a composite object $a$ is constructed by recursively applying $\Theta$ to fuse from fundamental objects as in \eqnref{eq: associativity}.

\begin{figure}[htbp]
\begin{center}
\includegraphics[width=\columnwidth]{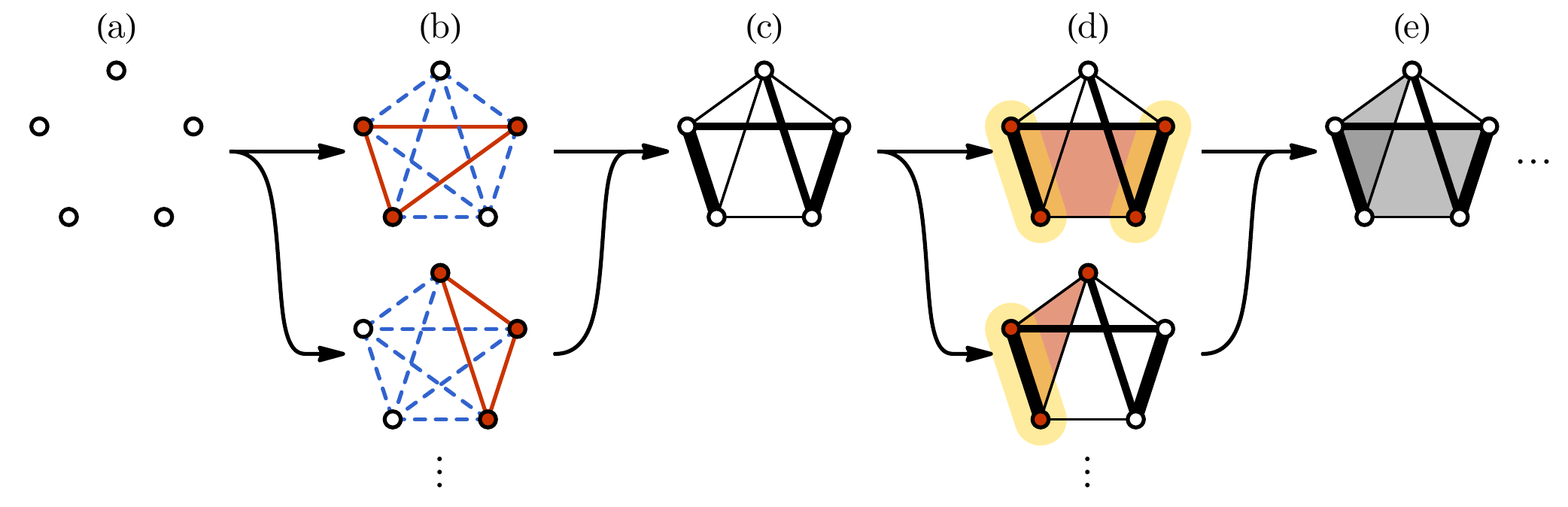}
\caption{Bootstrap approach for multi-scale categorical representation learning. (a) Starting with a set of elementary objects. (b) Sample concurrent objects from the data set (high-lighted as red nodes). Strengthen the connection among the concurrent objects (red links) and weaken other links. (c) After training, the model learns about the relationships of different strengths. (d) Move to the next level by fusing pairs of objects with strong connections  (covered in yellow shades) to form compound objects. Sample concurrent objects from the data set to train the relationships among compound and elementary objects. (e) The model learns higher relations.}
\label{fig: RG}
\end{center}
\end{figure}

A more challenging situation is that the data set has no naturally-defined levels of scopes, such that the hierarchical representation must be established with a bootstrap approach. We assume that at least a set of elementary objects can be specified in the data set (such as words in the language data set), illustrated as small circles in \figref{fig: RG}(a). We first apply the categorical representation learning approach to learn the linking probability $P(a\to b)$ between objects. The algorithm first samples different clusters of objects from the data set within a certain scale. The linking probability is enhanced for the pair of objects concurring in the same cluster, and is suppressed otherwise, as shown in \figref{fig: RG}(b). In this way, the algorithm learns to find the embedding $v_a$ for every fundamental object $a$. After a few rounds of training, fuzzy morphisms among objects will be established, as depicted in \figref{fig: RG}(c). The thicker link represents higher linking probability $P(a\to b)$ and stronger relations. In later rounds of training, when the sampled cluster contains a pair of objects $(a,b)$ connected by a strong relation (with $P(a\to b)$ beyond a certain threshold), it will be considered as a composite object $a\otimes b$, as yellow groups in \figref{fig: RG}(d). The embedding $v_{a\otimes b}$ of composite object will be calculated from that of the fundamental objects as $v_{a\otimes b}=\Theta(v_a\otimes v_b)$, based on the fusion operator $\Theta$ given by the tensor bifunctor model. The model will continue to learn the linking probability $P(a\otimes b\to c)$ between the composite object $a\otimes b$ and the other object $c$ in the cluster, as red polygons in \figref{fig: RG}(d). In this way, the fusion operator $\Theta$ will get trained together with the object and morphism embeddings. This will establish higher morphisms between composite objects like \figref{fig: RG}(e). The approach can then be carried on progressively to higher levels, which will eventually enable the machine to learn the hierarchical structures in the data set and establish the representations for objects, morphisms and tensor bifunctors all under the same approach.

\begin{appendix}

\section{Preliminary results}
\subsection{Learning Chemical Compounds}
\subsubsection*{Data Set}
\begin{figure}[htbp]
\begin{center}
\includegraphics[width=\columnwidth]{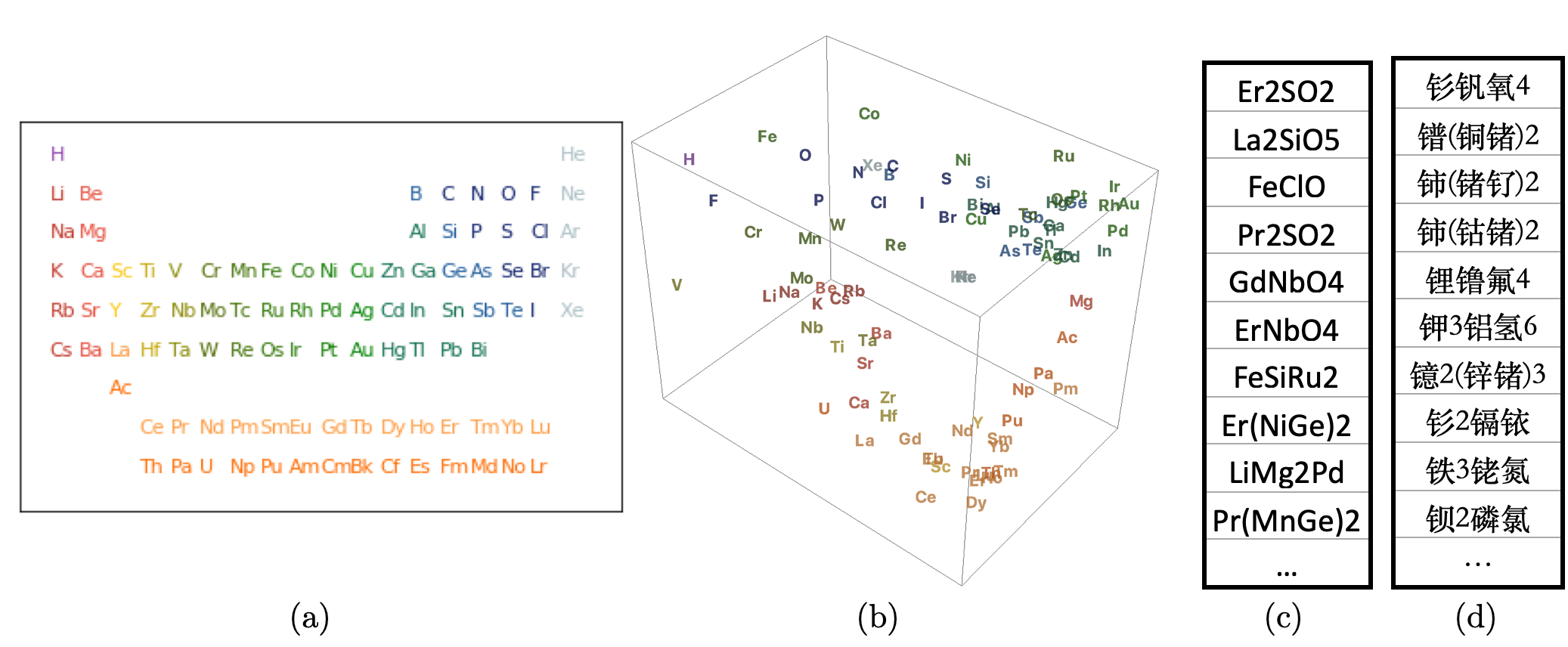}
\caption{(a) Periodic table. (b) Embeddings of elements by singular value decomposition of the point-wise mutual information. (c) Examples of compounds (in English). (d) Examples of compounds (in Chinese).}
\label{fig: dataset}
\end{center}
\end{figure}

To demonstrate the proposed categorical learning framework, we apply our approach to the inorganic chemical compound data set. The data set for our proof of concept (POC) contains 61023 inorganic compounds, covering 89 elements in the periodic table \figref{fig: dataset}(a). The data set can be modeled as a category, which contains elements as fundamental objects, as well as functional groups and compounds as composite objects. The morphisms represent the relations (such as chemical bonds) between atoms or groups of atoms. We assume that the concurrence of two elements in a compound is due to the underlying relations, such that the morphisms can emerge from learning the elements' concurrence.

On the data set level, we  collect the point-wise mutual information (PMI) $\mathsf{PMI}(a,b)=\log p(a,b) -\log p(a)-\log p(b)$, where $p(a,b)$ is the probability for the pair of elements $a$, $b$ to appear in the same chemical compound, and $p(a)$ is the marginal distribution. Performing a principal component analysis of the PMI and taking the leading three principal components, we can obtain three-dimensional vector encodings of elements, as shown in \figref{fig: dataset}(b). We observe that elements of similar chemical properties are close to each other, because they share similar context in the compound. This observation indicates that our algorithm is likely to uncover such relations among elements from the data set.

\subsection{Unsupervised/Semisupervised Translation}

To demonstrate the categorical representation learning and the functorial learning in sections 1 and 2, we designed an unsupervised translation task with the chemical compound data set. We take the data set in English and translate each element into Chinese. \figref{fig: dataset}(c,d) shows some samples from both the English and Chinese data set. The task of unsupervised (or semisupervised) translation is to learn to translate chemical compounds from one language to another without aligned samples (or with only a few aligned samples). The unsupervised translation is possible since the chemical relation between elements are identical in both languages. The categorical representation learning can capture these relations and represent them as morphism embeddings. By aligning the morphism embeddings, using the funtorial learning approach, the translator can be learned as a functor that maps between the English and Chinese compound categories. The translation functor is required to map elements to elements while preserving their relations. 

We demonstrate the semisupervised translation. We assign 15 elements as supervised data and provide the aligned English-Chinese element pairs to the machine. By minimizing the structure and alignment loss together. 

\begin{figure}[htbp]
\begin{center}
\includegraphics[width=0.75\columnwidth]{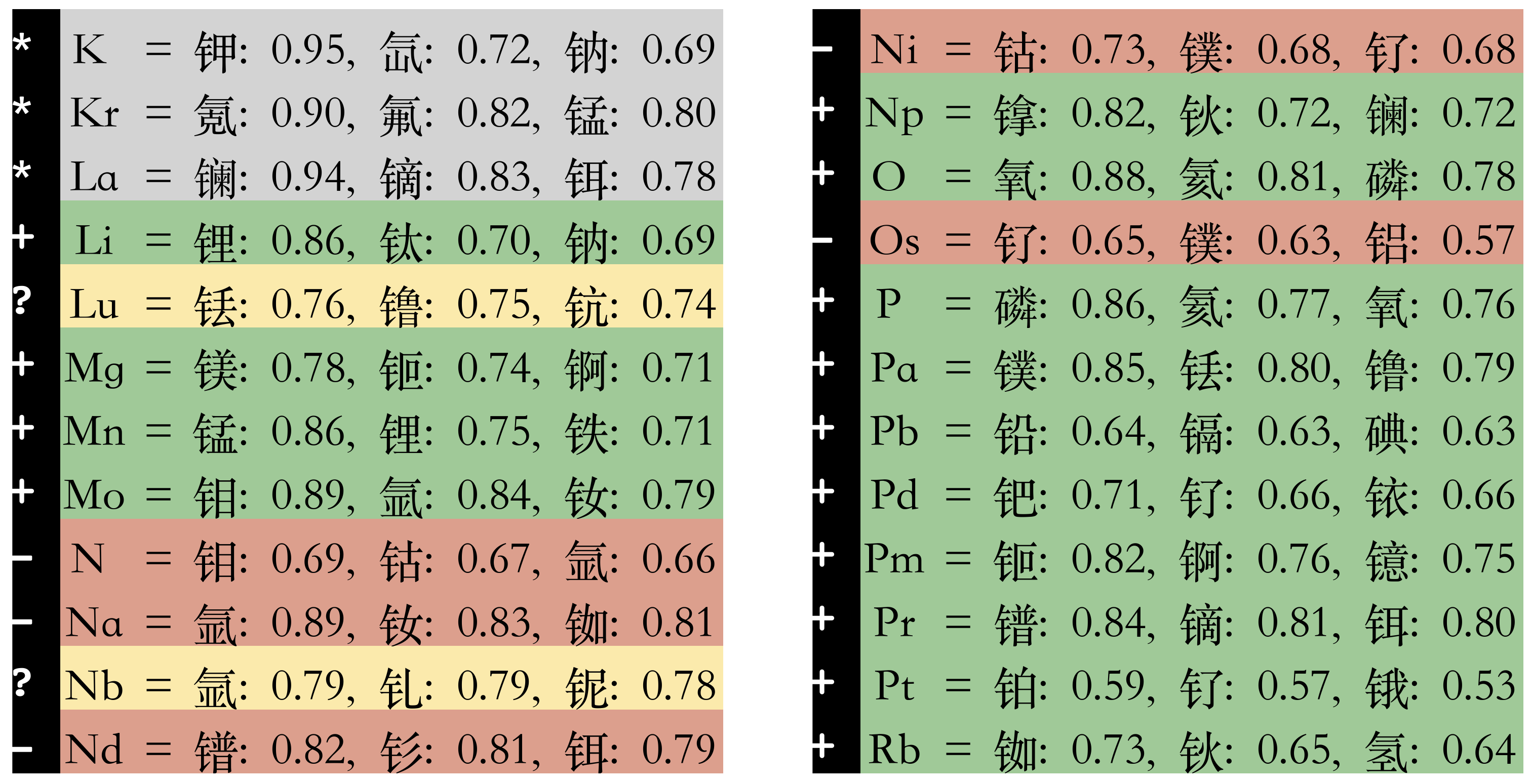}
\caption{Results of the semisupervised translation. For each English element, the top three Chinese translations are listed. The gray lines are selected supervised elements. The row is green if the correct translation is the top candidate. The row is yellow if the correct translation is not the top candidate but appears within top three. The row is red if the correct translation does not appear even within the top-three candidates.}
\label{fig: results}
\end{center}
\end{figure}

\subsection{More on model structure for translator (by Ahmadreza Azizi)}
\begin{figure}[h]
    \centering
    \includegraphics[width=0.8\linewidth]{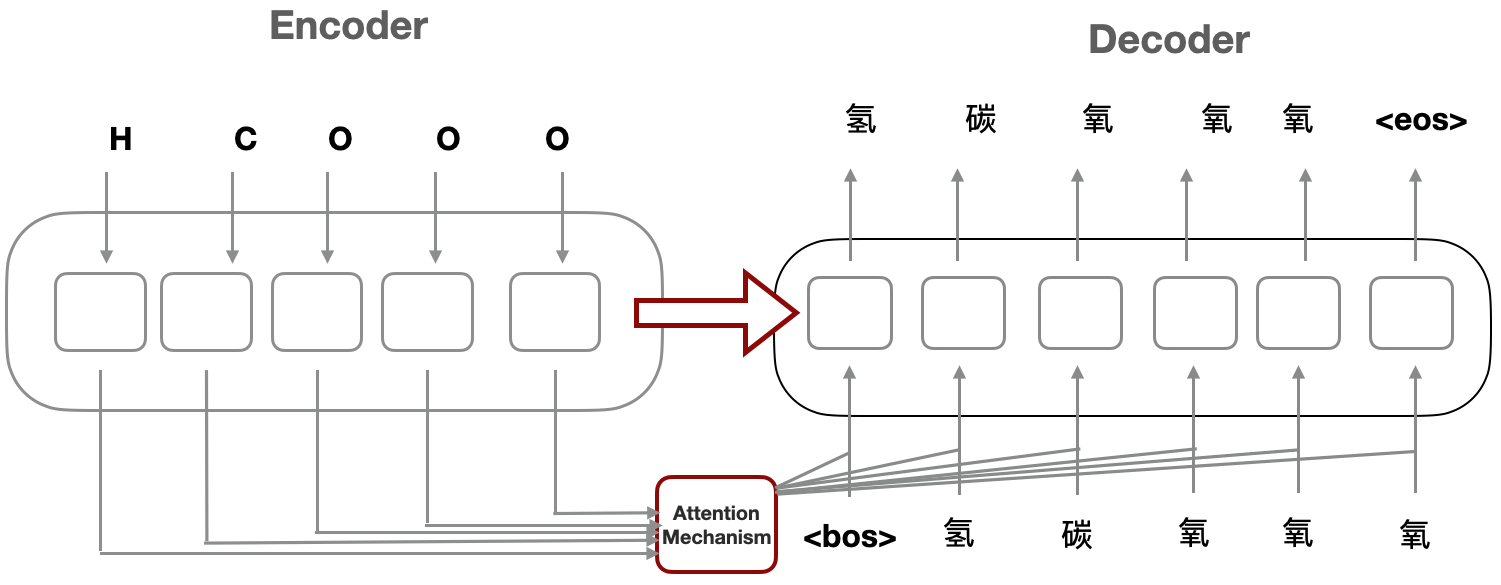}
    \caption{Structure of deep learning models with GRU cells.}
    \label{fig:seq2seq}
\end{figure}

 Many of the state-of-the-art translation models (usually called sequence to sequence or seq-2-seq models) incorporate two blocks of Encoder and Decoder that are connected to the source input and the target input respectively. Also in different models the Encoder and Decoder blocks are connected to each other in different ways. For the purpose of POC, we hence use the most relevant design in the seq-2-seq models and compare its results with our categorical learning model. Recently, text translation models with transformers have had impressive achievements on the datasets with long inputs (usually more than 250 tokens). However we do not use transformers, due to the name of compounds not being long and hence transformers do not have specific supremacy over their counterparts with recurrent neural network (RNN) based cells in this case. Therefore our choice is to design a seq-2-seq model which includes two blocks of Encoder and Decoder with general recurrent unit (GRU) cells which benefits from the attention mechanism (see Figure \ref{fig:seq2seq}). We add attention mechanism to the model as there is no significant language pattern in the compounds, hence the connection between Encoder and Decoder must be empowered with attention mechanism so that almost all the information in the Encoder can be expressed to the Decoder. Finally, we apply the teacher-force technique to provide aid to the Decoder for translating the English name of compounds to the Chinese counterparts, more robustly.\\

\subsubsection{Learning method for translator}
Formally, the source sentence $s$ is a set of words $s=\{s_1,s_2,...,s_T\}$ and the target sentence is $t=\{t_1,t_2,...,t_T\}$ both with length $T$. The embedding layers of source sentence and target sentence represent them as two sets of vectors $x=\{x_1,x_2,..,x_T\} \in X$ and $y=\{y_1,y_2,..,y_T\} \in Y$ respectively. For each element in $x$, the Encoder $E$ reads $x_t$ and encapsulates its information in the hidden states 
$$h_{t}=E(h_{t-1},x_{t-1}).$$
After reading all elements in $x$, the Encoder outputs the last hidden layer $h_{T}=E(h_{T-1},x_{T-1})$ to be passed to the first cell in Decoder. Therefore for each element $y_{t}$ in $y$, the hidden state of the decoder cell at $t$ is given by 
$$s_{t}=D(s_{t-1}, y_{y},c_{t}).$$
In this formula, the attention mechanism is expressed as the context vector $c_{t}$. Conceptually, the context vector  $c_{t}=\sum_{i=1}^{T}\alpha_{t,i} h_{i}$ carries out the sum of information of the encoder cells ($h_i$) weighted by the alignment score $\alpha_{t,i}$. Here the score function $\alpha_{t,i}$ indicates how much information from each input $x_i$ should be contributed to the output $y_t$ and can have different forms \cite{bahdanau2014neural,vaswani2017attention}.
\\
Finally, given the output of Decoder cells $ \hat{y}$, its similarity will be compared to the true translation $y$ thorough the cross entropy loss function:
\begin{equation}
    L(\theta_D,\theta_E)=-\frac{1}{N} \sum_{i} y_i \ log \ \hat{y_i}
    \label{eq::L}
\end{equation}
with $\theta_E$ and $\theta_D$ the trainable parameters in the Encoder block and the Decoder block respectively. The Equation \ref{eq::L} can be seen as a metric function that measures the distance between $y$ and $\hat{y}$.

\subsubsection{Translator results}
According to Equation \ref{eq::L}, in the process of training, the parameters $\Theta_{E}$ and $\Theta_{D}$ are optimized so that the model output $\hat{y}$ becomes very similar to the ground truth $y$. We train the deep learning model with various sizes of GRU cells and count number of correct translations that are made in each model. 
\begin{table}[h]
    \renewcommand{\arraystretch}{1}
    \setlength{\extrarowheight}{0.2cm}
    \centering
    \begin{tabular}{|>{\centering} p{4.5cm}| >{\centering} p{2.5cm}|>{\centering} p{1.5cm}|>{\centering} p{1.5cm}|>{\centering\arraybackslash} p{1.5cm}|}
    \hline
    Models &  Categorical Learning & Seq2Seq & Seq2Seq & Seq2Seq \\[0.5cm] 
    \hline
     Number of parameters & 16,600 & 25,626 & 77,222 & 287,020  \\[0.2cm]
    \hline
    Number of supervised elements &  15 & 15 & 15  & 15
    \\[0.2cm]
    \hline
    Number of correct translations & 57 & 12 & 38  & 59 \\
    \hline    
    \end{tabular}
    \caption{Performance of deep learning models and categorical learning model after training on compounds data.}
    \label{tab::results}
\end{table}

Note that the categorical learning model has only 16600 parameters and we believe it is only fair if only we compare the deep learning model with similar number of parameters. Table \ref{tab::results} elaborates on our results for deep learning models and our categorical learning model. These results indicate that the deep learning models are unable to outperform our model unless the number of parameters they use is 17 times more than our categorical learning model, in which case the deep learning model has the capability of achieving a better performance. This comparison is particularly interesting, once one decreases the number of parameters to about 25,000. Our experiments clearly indicate that the categorical learning methods completely outperform the deep learning models. 

\end{appendix}

\newpage
\bibliographystyle{unsrturl}
\bibliography{ref}

\noindent{\small{\tt{artan@cmsa.fas.harvard.edu, yzyou@physics.ucsd.edu}}

\end{document}